# Deep Scattering: Rendering Atmospheric Clouds with Radiance-Predicting Neural Networks

SIMON KALLWEIT, Disney Research and ETH Zürich
THOMAS MÜLLER, Disney Research and ETH Zürich
BRIAN MCWILLIAMS, Disney Research
MARKUS GROSS, Disney Research and ETH Zürich
JAN NOVÁK, Disney Research

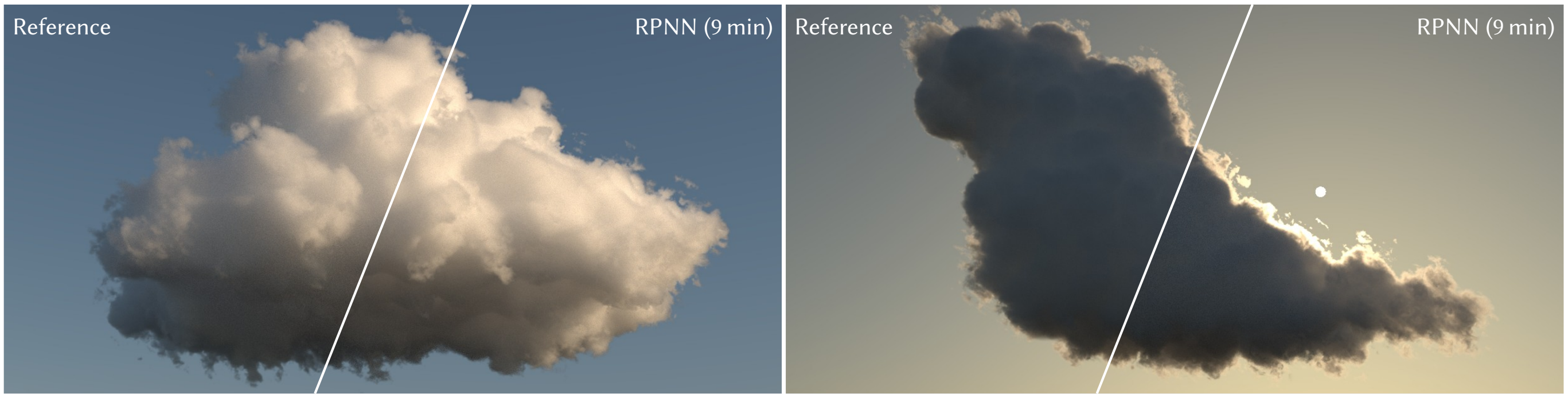

Fig. 1. We synthesize multi-scattered illumination in clouds using deep radiance-predicting neural networks (RPNN). We combine Monte Carlo integration with data-driven radiance predictions, accurately reproducing edge-darkening effects (left), silverlining (right), and the whiteness of the inner part of the cloud.

We present a technique for efficiently synthesizing images of atmospheric clouds using a combination of Monte Carlo integration and neural networks. The intricacies of Lorenz-Mie scattering and the high albedo of cloud-forming aerosols make rendering of clouds—e.g. the characteristic silverlining and the "whiteness" of the inner body—challenging for methods based solely on Monte Carlo integration or diffusion theory. We approach the problem differently. Instead of simulating all light transport during rendering, we pre-learn the spatial and directional distribution of radiant flux from tens of cloud exemplars. To render a new scene, we sample visible points of the cloud and, for each, extract a hierarchical 3D descriptor of the cloud geometry with respect to the shading location and the light source. The descriptor is input to a deep neural network that predicts the radiance function for each shading configuration. We make the key observation that progressively feeding the hierarchical descriptor into the network enhances the network's ability to learn faster and predict with higher accuracy while using fewer coefficients. We also employ a block design with residual connections to further improve performance. A GPU implementation of our method synthesizes images of clouds that are nearly indistinguishable from the reference solution within seconds to minutes. Our method thus represents a viable solution for applications such as cloud design and, thanks to its temporal stability, for high-quality production of animated content.

## 1 INTRODUCTION

Efficient and accurate rendering of high-albedo materials is a challenging problem, especially when the optical properties vary spatially. Heterogeneous densities of aerosols formed into visible structures, e.g. atmospheric clouds, is one such class of materials that challenge the efficiency of rendering algorithms. The high reflectivity of liquid droplets and frozen crystals produces appearances dominated by multiple scattering of light, often on the order of thousands of photon-matter interactions. Even if the discrete nature is ignored and the cloud is approximated by a continuous volume, estimating light transport by solving the radiative transfer equation (RTE) [Chandrasekhar 1960] is computationally intensive.

Many numerical recipes for synthesizing photorealistic images are based on Monte Carlo (MC) integration. While these techniques have been successfully applied to volume rendering, and despite the theory naturally extends to multi-dimensional problems, MC rendering of atmospheric clouds faces two challenges. Firstly, the space of paths that transport non-negligible contributions is *many*-dimensional, requiring tremendous amounts of samples to reduce estimation variance. Secondly, the construction of individual path

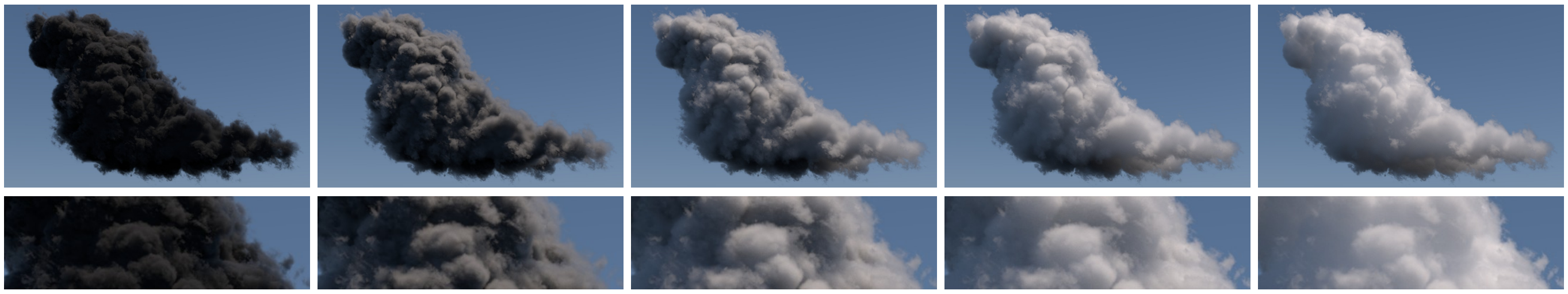

(a) 1 interaction
Render time = 0.017 h

(a) 4 interactions
Render time = 0.89 h

(a) 16 interactions
Render time = 6.5 h

(a) 64 interactions
Render time = 16 h

(a) All interactions
Render time = 34 h

Fig. 2. Multiply-scattered light transport is paramount for synthesis of realistic images of clouds. We show renderings of the same cloud under equal illumination where up to 1, 4, 16, 64, and arbitrarily many light-matter interactions are considered per light path. Because the albedo of clouds is close to 1, very long chains of light-matter interactions still carry significant amounts of energy and thus need to be considered, resulting in high render times. For each configuration we report the render time it takes to converge to the variance of the right-most image.

samples is expensive even with sophisticated techniques for free-path sampling [Kutz et al. 2017; Szirmay-Kalos et al. 2017] and advanced acceleration data structures [Szirmay-Kalos et al. 2011; Yue et al. 2010]. Rendering realistic clouds with path-tracing algorithms thus requires tens of hours of computation to obtain images with acceptable error; see Figure 2.

While such cost may be acceptable in some scenarios, interactive and throughput-oriented applications, such as asset design and movie production, require frames to be computed orders of magnitude faster. These typically employ approximations based on diffusion theory [Koerner et al. 2014; Stam 1995], density estimation [Elek et al. 2012], or semi-analytic solutions [Bouthors et al. 2008] heavily reducing visual quality in favor of fast rendering. We similarly trade accuracy for higher speed, but tackle the problem by synthesizing high-order scattering using neural networks.

Deep neural networks consist of compositions of simple building blocks that, when combined in the right way, achieve state-of-the-art performance in a wide variety of computational domains [LeCun et al. 2015]. Their application to transport problems is however still relatively unexplored. We demonstrate that, when designed carefully, these networks can accurately predict the distribution of radiance in clouds due to unoccluded distant lighting upon arbitrarily many scattering events.

Since light propagation is a non-local phenomenon, the prediction speed hinges on the ability to concisely express the cloud geometry that is relevant to a given shading configuration. To that end, we employ a hierarchical point stencil that accurately captures the local neighborhood and compactly approximates remote regions and the overall bulk of the cloud. We observe that feeding the stencil to the network *all-at-once* yields poor results. Instead, we propose to input the hierarchy progressively, with each block of the network having direct access to at most one level of the hierarchy. Such design drives the network to reflect on the multi-scale nature of light transport and reduces the number of trainable parameters. This is key for performing many fast predictions needed to render the image.

We compare our method to reference path-traced images and the state of the art in grid-resolved diffusion theory. Our results are significantly closer to reference images and render as fast as the other approximate method.

## 2 RELATED WORK

*Monte Carlo Methods.* Kajiya and Von Herzen [1984] were the first to use path tracing for numerically estimating radiative transfer in volumes [Chandrasekhar 1960]. Their technique was later extended by constructing paths in a bidirectional manner [Lafortune and Willems 1996], mutating paths using the Metropolis-Hastings algorithm [Pauly et al. 2000], and importance sampling of low-order scattering [Georgiev et al. 2013; Kulla and Fajardo 2012]. The efficiency of these algorithms can be increased by reusing computation and correlating estimates as in radiance caching [Jarosz et al. 2008], many-light rendering [Novák et al. 2012; Raab et al. 2008] and density-estimation-based methods [Jarosz et al. 2011, 2008; Jensen and Christensen 1998]. Křivánek et al. [2014] formulated a unified theory for these seemingly incompatible approaches, allowing to combine their strengths in a robust estimator. Bitterli and Jarosz [2017] generalized volumetric density estimation to arbitrary-dimensional samples. Nevertheless, even with free paths and transmittance estimates constructed using advanced approaches [Kutz et al. 2017; Novák et al. 2014; Szirmay-Kalos et al. 2017] operating on special data structures [Szirmay-Kalos et al. 2011; Yue et al. 2011, 2010], these methods are far from reaching interactive frame rates when used on the highly scattering materials that we target.

*Diffusion Theory.* Certain situations permit approximating multiple scattering with the diffusion theory, in graphics pioneered by Stam [1995] who rendered heterogeneous media by solving a discretized form of the diffusion equation on a grid. Koerner et al. [2014] modulate the diffusion coefficient to increase accuracy in regions with low density and/or high albedo. Several semi-analytic solutions based on combining transport contributions from two or more monopoles were developed for fast simulation of subsurface scattering [d'Eon and Irving 2011; Donner and Jensen 2005; Frisvad et al. 2014; Jensen et al. 2001]. Unfortunately, none of the diffusion approaches handle the highly anisotropic scattering in clouds well, and the appearance stemming from low- and mid-order scattering, e.g. silverlining, is not reproduced well.

*Tabulated Multiple Scattering.* The anisotropy can be better preserved when using precomputation-based solutions. Szirmay-Kalos et al. [2005] trace a number of light paths to build illumination networks and reevaluate the transport on the fly achieving real-time

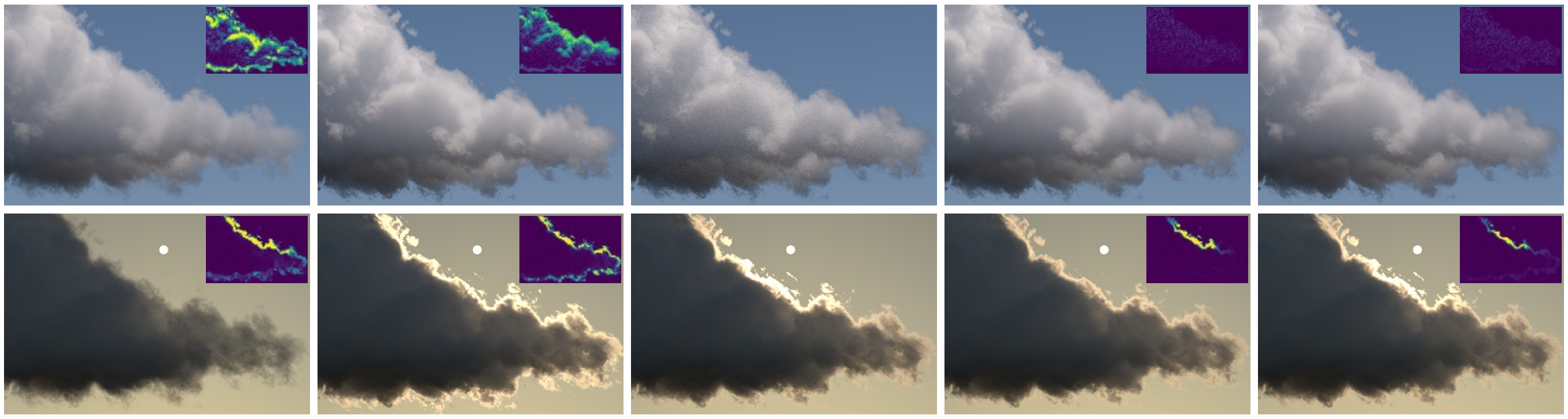

(a) Isotropic phase function Equal-time variance = 17.7 (b) Henyey-Greenstein, $g = 0.857$ Equal-time variance = 698 (c) Lorenz-Mie (LM) Equal-time variance = 23136 (d) Chopped LM Equal-time variance = 136 (e) Chopped LM after 1 bounce Equal-time variance = 144 frame rates. Moon et al. [2007] proposed to tabulate scattered radiance on spherical shells surrounding a directional point emitter embedded in a granular material. An analogous approach was used by Lee and O'Sullivan [2007] to handle homogeneous continuous volumes, and further adapted to heterogeneous media by Müller et al. [2016]. All these methods are either interactive, or produce high-fidelity images, but none of them achieve both concurrently.

Fig. 3. Front-lit (top) and back-lit (bottom) cloud from Figure 2 rendered with different phase functions. Isotropic scattering (a) *reduces* contrast and fails to synthesize silverlining. The Henyey-Greenstein approximation (b) incorrectly *increases* contrast (top) and produces a different silverlining effect than Lorenz-Mie (LM) scattering (c). Chopping the peak (d) leads to lower estimation variance, but also removes the strong silverlining. Using the complete LM phase function at the first bounce only and then switching to the chopped version provides nearly perfect result and low estimation variance (e).

*Specialized Methods for Rendering Clouds.* Fast modeling and rendering of clouds was first addressed by Gardner [1985] who used ellipsoids with spatially varying densities obtained as a superposition of sine waves. Real-time rates can be achieved on graphics hardware by rasterizing particles [Harris and Lastra 2001], metaballs [Dobashi et al. 2000], or volume slices [Riley et al. 2004] into billboard textures and projecting them onto the image plane. At best, these approaches capture only a limited amount of (anisotropic) scattering; backscattering is often completely ignored. Neglecting high-order scattering, as proposed by Nishita et al. [1996], leads to poor results (according to current standards) in optically thick clouds; see Figure 2.

Bouthors et al. [2008] acknowledge the anisotropy by employing Lorenz-Mie scattering. We also adopt Lorenz-Mie scattering with the assumption of spherical scatters; non-spherical scatterers have been studied by Frisvad et al. [2007].

Bouthors et al. [2008] further propose to precompute solutions for flat slabs that are then fitted into the cloud geometry achieving real-time frame rates. Instead of slabs, Elek et al. [2012] record photons in a regular grid where each cell models the angular distribution using the Henyey-Greenstein (HG) basis. In a follow-up [Elek et al. 2014], the authors increased the performance by combining the HG basis with the discrete-ordinates method. Nevertheless, for albedos close to 1 the approach requires either excessive numbers of iterations or multiple super-imposed low-resolution grids that tend to blur high-frequency illumination patterns.

Our goal is to accurately synthesize all-order, anisotropic scattering and accurately capture the appearance of the dense inner part and wispy boundaries. We explore a new approach based on approximating the cloud geometry by a hierarchical descriptor and predict local illumination using a deep neural network.

*Neural Networks.* Deep neural networks (see Bengio et al. [2013]; LeCun et al. [2015] for a comprehensive review) are able to efficiently model complex relationships between input and output variables in a highly non-linear manner. This data-driven approach has emerged as a state-of-the-art technology in a wide variety of challenging problems, e.g. image recognition [He et al. 2016; Simonyan and Zisserman 2014], machine translation [Wu et al. 2016], or generative modeling of raw audio and natural images [Oord et al. 2016a,b].

Deep learning has also been successfully applied to problems in computer graphics. Nalbach et al. [2017] use convolutional neural networks (CNNs) to synthesize ambient occlusion, illumination, and other effects in screen space. Bako et al. [2017] and Chaitanya et al. [2017] employ neural networks for denoising rendered images. These approaches operate in 2D, using color and feature images as inputs to the network. Similarly to Chu and Thuerey [2017], who apply CNNs to fluid simulation, our networks operate in 3D.

Closely related to ours is the work by Ren et al. [2013], who train 2-hidden-layer MLPs to predict radiance in a specific region of a scene. In contrast, our 23-layer MLPs are applicable to an entire spectrum of atmospheric clouds. We use a hierarchical feature and feed its levels into the network progressively. While the first provides a cost-effective extraction of global information, the latter keeps the number of trainable parameters low facilitating fast predictions.

## 3 RADIATIVE TRANSFER IN CLOUDS

In this section, we discuss properties of clouds that are key for synthesizing their appearance correctly (see Figure 3). We then briefly review the RTE to allow defining the quantity to be predicted and describing our approach in the following section.

### 3.1 Properties of Clouds

Clouds can be primarily categorized as a cirrus (thin, wispy curls), a stratus (hazy, featureless sheets), or a cumulus (fluffy heaps). The last type is the most challenging to render due to its high-frequency boundary details and the smooth bright appearance of the inner body stemming from many interactions that light undergoes inside.

The aerosols forming a cloud consist of many small water particles—droplets or ice crystals—the placement and size of which can be well represented statistically. The radii of droplets in a cumulus cloud range from 1 to 100 microns with means typically around 5 $\mu$ [Weickmann and Kampe 1953]. The droplet-size distribution function (DSD) has a key impact on the angular distribution of scattered light, and can be well approximated by a gamma function. Because these droplets absorb little light, long chains of light-matter interactions are an integral part of the characteristic appearance of clouds (see Figure 2). Since the scatterers are comparable in size to the wavelength of visible light, the distribution of scattered light, represented by phase function $p$, needs to be derived from Maxwell's equations. Lorenz [1890] and Mie [1908] developed solutions for planar waves interacting with a homogeneous set of spheres. The resulting phase function exhibits a very strong diffraction peak, a wide forward-scattering lobe, a back-scattering circular ridge, and a back-scattering peak; see Figure 4 for an example and the supplementary material for details.

As shown in Figure 3, the intricate shape of the Lorenz-Mie phase function produces characteristic visual effects such as silverlining, fogbow, and glory, which are lost or synthesized inaccurately when using an isotropic or Henyey-Greenstein approximation. At the same time, however, using the high-frequency, multi-modal Lorenz-Mie distribution makes it difficult to sample the high-energy, multi-scattered transport efficiently; too difficult to render our training data within reasonable time. We follow the suggestion of Bouthors et al. [2008] to "chop" the diffraction peak and lower the cloud density according to the fraction of scattered light contained in the peak. The explicit simulation of near-perfect forward scattering—the main source of sampling difficulties—is thus removed and accounted for implicitly by reducing the optical thickness; see the supplementary material for details.

As shown in Figure 3(d), the chopped Lorenz-Mie reduces estimation variance and produces accurate results in front-lit or side-lit configurations. However, since the slight angular blurring due to the diffraction peak is replaced by continuing in the perfectly forward direction, the very prominent effect of silverlining in backlit configurations is nearly completely lost. Since the silverlining stems from low-order scattering, we opt to use the full Lorenz-Mie phase function for the *first* bounce and then switch to the chopped version and its corresponding optically thinner volume. This preserves the effect of silverlining and provides sufficient performance for creating a large set of ground-truth simulations to train our networks.

### 3.2 Radiative Transfer

As in other works, we approximate the discrete aerosol mixture by a continuous medium, parameterized by a spatially varying extinction coefficient $\mu_t(\mathbf{x})$, a globally constant albedo $\alpha$, and an axially symmetric phase function $p(\cos\theta)$. In such a medium, the change

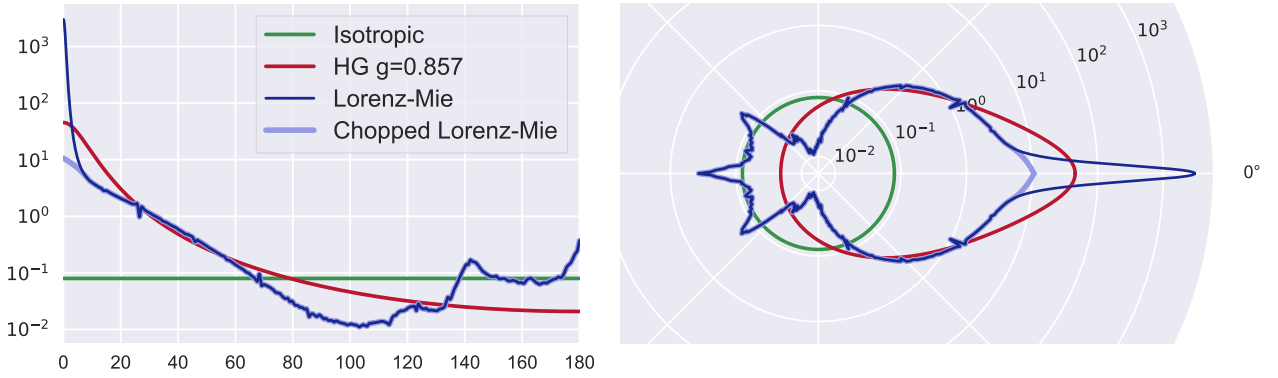

Fig. 4. Isotropic (green), Henyey-Greenstein with $g = 0.857$ (red), and Lorenz-Mie (dark blue) phase functions. The light blue shows an unnormalized version of Lorenz-Mie with a chopped diffraction peak.

in radiance $L$ at point $\mathbf{x}$ in direction $\omega$ is governed by the radiative transfer equation:

$$(\omega \cdot \nabla)L(\mathbf{x},\omega) = -\mu_t(\mathbf{x})L(\mathbf{x},\omega) + \mu_s(\mathbf{x})\int_{S^2} p(\omega\cdot\widehat{\omega})L(\mathbf{x},\widehat{\omega})\,\mathrm{d}\widehat{\omega},$$

Integrating both sides of the differential RTE along $\omega$ yields the following Fredholm integral equation of the second kind:

$$L(\mathbf{x},\omega) = \int_0^\infty \exp\left(-\int_0^u \mu_t(\mathbf{x}_v)\mathrm{d}v\right)\mu_s(\mathbf{x}_u)\int_{S^2} p(\omega\cdot\widehat{\omega})L(\mathbf{x}_u,\widehat{\omega})\mathrm{d}\widehat{\omega}\,\mathrm{d}u,$$

where $\mathbf{x}_u = \mathbf{x} - u\omega$ and analogously for $\mathbf{x}_v$. The exponential term represents the attenuation of light on a straight line between $\mathbf{x}$ and $\mathbf{x}_u$ commonly referred to as transmittance $T(\mathbf{x},\mathbf{x}_u)$.

Since clouds are non-emissive and light can only be injected from the outside, we need to define a proper boundary condition. Assuming $\mathbf{x}_b$ is on a hypothetical boundary $\Omega$ that encloses the cloud, i.e. $\forall \mathbf{x}_b \in \Omega : \mu_t(\mathbf{x}_b) = 0$, we have

$$L(\mathbf{x},\omega) = \int_0^b T(\mathbf{x},\mathbf{x}_u)\mu_s(\mathbf{x}_u)\int_{S^2} p(\omega\cdot\widehat{\omega})L(\mathbf{x}_u,\widehat{\omega})\,\mathrm{d}\widehat{\omega}\,\mathrm{d}u + T(\mathbf{x},\mathbf{x}_b)L(\mathbf{x}_b,\omega), \tag{1}$$

where $b$ is the distance to the boundary.

Equation (1) can be estimated by recursive Monte Carlo integration: we alternate sampling of distances and directions, gradually building a path towards the boundary and evaluating the radiance thereof (e.g. by extending the path beyond $\Omega$). Even with the latest importance-sampling techniques for building volumetric paths, rendering high-albedo clouds by path tracing is too costly for interactive applications and we use it only for constructing the training set.

## 4 RADIANCE-PREDICTING NEURAL NETWORKS

We propose a supervised learning approach to sidestep the costly evaluation of the recursion in Equation (1). Our goal is to efficiently synthesize the in-scattered radiance

$$L_s(\mathbf{x},\omega) = \int_{S^2} p(\omega\cdot\widehat{\omega})L(\mathbf{x},\widehat{\omega})\,\mathrm{d}\widehat{\omega}.$$

To that end, we considered learning and predicting the complete $L_s(\mathbf{x},\omega)$, but observed that the network struggles with predicting uncollided radiance, $L_d(\mathbf{x},\omega)$; i.e. the light that arrives from light sources directly and contributes to single scattering along camera rays. We thus opt for predicting only multi-scattered transport, precisely the *indirect* in-scattered radiance:

$$L_i(\mathbf{x},\omega) = \int_{S^2} p(\omega\cdot\widehat{\omega})(L(\mathbf{x},\widehat{\omega}) - L_d(\mathbf{x},\widehat{\omega}))\,\mathrm{d}\widehat{\omega}.$$

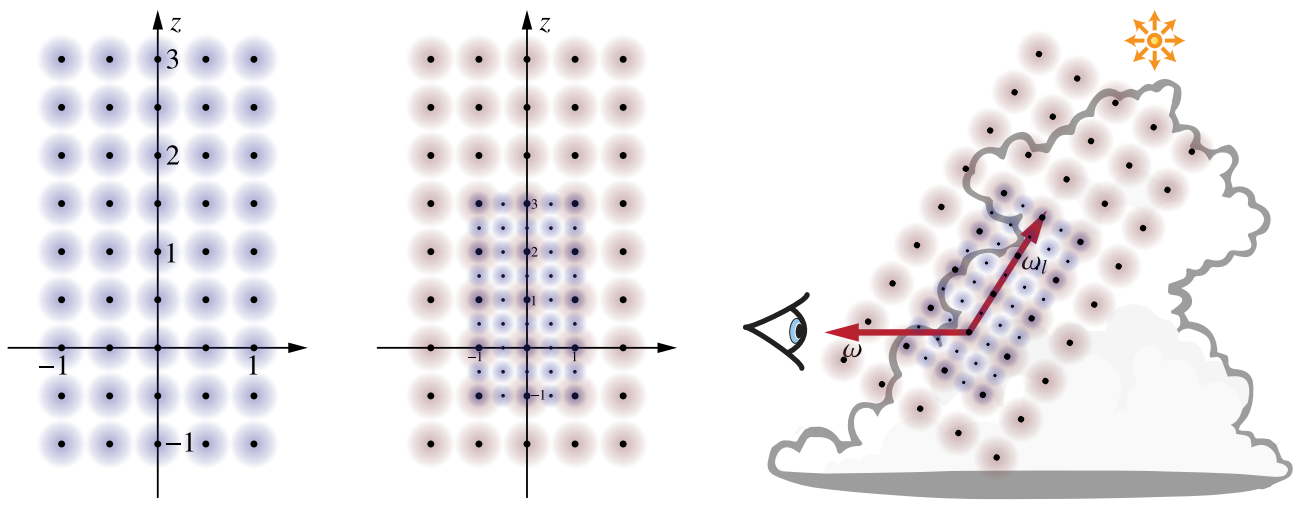

(a) Stencil grid (b) Two levels (c) Local frame

Fig. 5. The descriptor consists of $5 \times 5 \times 9$ stencils (a) that are progressively scaled and turned into a hierarchy (b), which is then oriented towards the light source (c); only two levels of the hierarchy are shown here (red and blue) to avoid clutter.

For the the rest of the transport we employ MC integration that samples visible points of the cloud (first integral in Equation (1)) and estimates single scattering thereof.

## 4.1 Learning from Data

The goal in supervised learning is to find a function $g(\mathbf{z}; \boldsymbol{\theta}) : \mathbb{R}^d \rightarrow \mathbb{R}$, which maps a feature descriptor $\mathbf{z}$ to a target value, $L_i(\mathbf{x}, \omega)$ in our case, and depends on a set of parameters, $\boldsymbol{\theta}$. Since light transport is global, $L_i(\mathbf{x}, \omega)$ is effectively a function of not only the location $\mathbf{x}$ and direction $\omega$, but also depends on the light source and the density structure of the entire cloud; we denote this set $\mathcal{S}$ and refer to it as "shading configuration". We define an efficient descriptor $\mathbf{z} = \phi(\mathcal{S}) \in \mathbb{R}^d$ to accurately represent $\mathcal{S}$ without excessive computation and storage requirements; details follow in Section 4.2.

Then, given a finite sample of labeled training data

$$\mathcal{D}_N = \{(\mathbf{z}_1, L_i(\mathbf{x}_1, \omega_1)), \dots, (\mathbf{z}_N, L_i(\mathbf{x}_N, \omega_N))\},$$

we aim to find the value of $\boldsymbol{\theta}$ that minimizes the average loss between the predictions and the target values,

$$\widehat{\boldsymbol{\theta}} \in \underset{\boldsymbol{\theta}}{\operatorname{argmin}} \frac{1}{N} \sum_{i=1}^{N} \ell(g(\mathbf{z}_i; \boldsymbol{\theta}), L_i(\mathbf{x}_i, \omega_i)). \quad (2)$$

We propose restricting $g$ to a particular neural-network architecture that we describe in Section 4.3.

## 4.2 Descriptor Construction

The main challenge of defining a good descriptor $\mathbf{z}$ for a given configuration $\mathcal{S}$ is in capturing relevant information from the density of the cloud. In theory, a neural network should be able to extract all necessary information from raw data. However, we demonstrate that the number of trainable parameters can be significantly reduced by providing the network with an easy-to-compute feature descriptor. Managing the complexity of the network in this way has the benefit of speeding up the training procedure. More importantly however, it speeds up the predictions; millions of which are needed to synthesize a single frame.

To represent small-scale detail near $\mathbf{x}$, but also the overall shape of the cloud, we employ a hierarchy of point stencils that sample the cloud density. The stencils at individual levels are identical, but their support increases geometrically with scaling factor $2^{k-1}$, where $k$ is the (1-based) index of the level; we use $K = 10$ levels in total. The stencil in the highest level covers a $\left(2^9\right)^3 \approx 10^8\times$ larger *volume* than the stencil at $k = 1$. In order to make the descriptor scale independent, we express the stencil dimensions and density values in terms of the mean free path; we do so by scaling the cloud geometry and density such that one unit of distance corresponds to *one* mean free path within the average density of the cloud; the average is computed over all non-zero voxels. We refer to the scaled density as $\rho$.

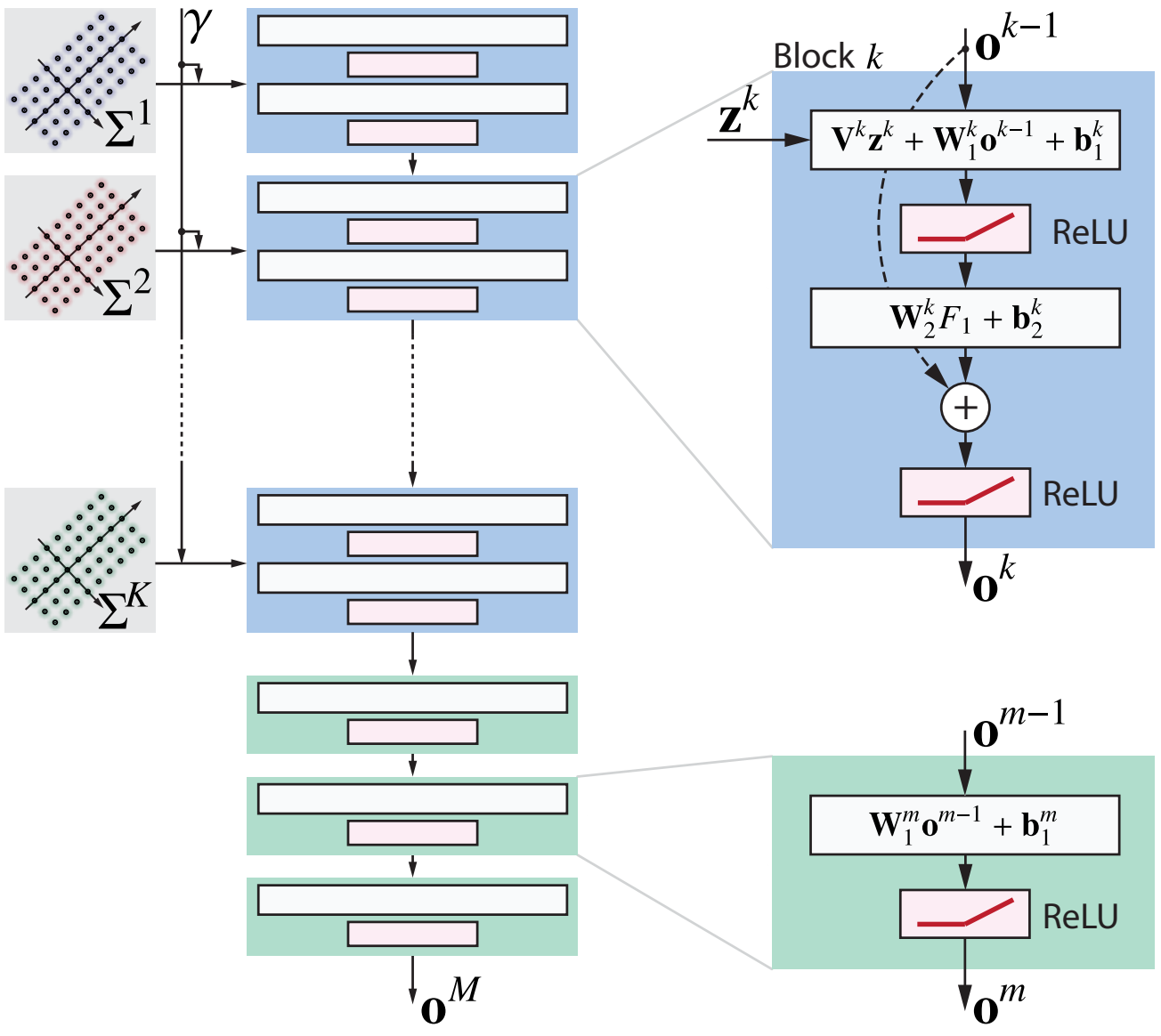

Fig. 6. We feed individual levels of the hierarchical stencil progressively. The upper part of the network consists of 2-layer blocks (blue) with residual connections, each of which processes the output of the previous block and one level of the stencil. We also add 3 standard layers (green) at the end. In all results we use $K = 10$, i.e. the complete network contains 23 layers.

All stencils share the same local frame centered at $\mathbf{x}$. The $z$-axis of the frame points towards the light source (e.g. the Sun or another distant emitter), the $x$-axis is perpendicular to the plane formed by the $z$-axis and $\omega$, and the $y$-axis is orthogonal to the two. The stencil at each level is formed by 225 points. Initially, these points form a $5 \times 5 \times 9$ grid in an axis-aligned box with $[-1, -1, -1]$ and $[1, 1, 3]$ being two opposing corners. The stencil at level $k$ is then further scaled by $2^{k-1}$; see Figure 5 for an illustration. For each level $k$, we extract the density values $\Sigma^k = \left\{\rho(\mathbf{q}_1^k), \rho(\mathbf{q}_2^k), \dots \rho(\mathbf{q}_{225}^k)\right\}$, the union of which, $\Sigma = \bigcup_{k=1}^{K} \Sigma^k$, represents the complete stencil.

The hierarchical stencil describes the cloud density with respect to location $\mathbf{x}$, direction towards the light source $\omega_l$, and the plane that $\omega_l$ forms with $\omega$. We also add the angle $\gamma = \cos^{-1}(\omega \cdot \omega_l)$ to complete the description of the configuration, i.e. $\mathbf{z} = \{\Sigma, \gamma\}$.

## 4.3 Network architecture

We use a network architecture based on a multilayer perceptron (MLP). In a standard MLP, each layer $m = 1, \dots, M$ applies a linear transformation to the output of the previous layer followed by an element-wise nonlinear activation function to obtain the layer output. We propose a modified architecture, tailored to the problem at hand, that yields better results than vanilla MLPs. We observe that the prediction quality improves if the individual levels of the stencil are fed into the network progressively, and that the performance further increases with residual connections [He et al. 2016].

The architecture is visualized in Figure 6. For each of the $K$ levels of the hierarchical stencil $\Sigma$, we construct a *block* structure that consists of two fully-connected layers which are bypassed using a residual connection. Each block $k$ processes the corresponding stencil level $\Sigma^k$ and angle $\gamma$, denoted $\mathbf{z}^k = \{\Sigma^k, \gamma\}$, and the output of the previous block $\mathbf{o}^{k-1}$ as follows

$$\begin{aligned} F_1^k &= f\left(\mathbf{V}^k \mathbf{z}^k + \mathbf{W}_1^k \mathbf{o}^{k-1} + \mathbf{b}_1^k\right), \\ F_2^k &= \mathbf{W}_2^k F_1^k + \mathbf{b}_2^k, \\ \mathbf{o}^k &= f\left(F_2^k + \mathbf{o}^{k-1}\right), \end{aligned}$$

where $\mathbf{V}$, $\mathbf{W}$ and $\mathbf{b}$ are trainable weights and biases, respectively. We use the rectified linear unit (ReLU) activation function $f(a) = \max(0, a)$. The residual connections serve two purposes: (i) they are necessary to train deeper networks efficiently [Balduzzi et al. 2017] and (ii) they encourage the network to learn how best to combine each layer in the input hierarchy with the information from previous levels in the stencil.

We also add a number of standard fully connected layers of the form $\mathbf{o}^m = f\left(\mathbf{W}_1^m \mathbf{o}^{m-1} + \mathbf{b}_1^m\right)$ to process the output of the last block. The output of the very final layer represents the predicted value, i.e. $g(\mathbf{z}; \boldsymbol{\theta}) = \mathbf{o}^M$, where $M$ is the total number of blocks and standard layers, and the set of parameters $\boldsymbol{\theta}$ consists of[1]:

$$\boldsymbol{\theta} = \{\mathbf{V}^k, \mathbf{W}_1^k, \mathbf{W}_2^k, \mathbf{b}_1^k, \mathbf{b}_2^k\}_{k=1}^K + \{\mathbf{W}_1^m, \mathbf{b}_1^m\}_{m=K+1}^M.$$

### 4.4 Rendering

In order to synthesize images using our network, we integrate it into a Monte Carlo ray tracer. We stochastically sample the first scattering interaction with delta tracking [Woodcock et al. 1965] and estimate *direct* in-scattering via Monte Carlo and predict *indirect* in-scattering with our neural network. Our renderer supports arbitrary distant light sources, which can be approximated well by drawing a number of directional samples. The direct in-scattering is estimated using next-event estimation, i.e. by combining samples from the phase function and the light source via multiple importance sampling. To predict indirect in-scattering, we extract descriptor $\mathbf{z}$ of the shading configuration and query our network for $L_i(\mathbf{x}, \omega)$. The direct and indirect components are added, multiplied by $\mu_s$ to obtain the absolute in-scattered radiance, and propagated back to the camera. By numerically integrating the predictions over the solid angle of the environment emitter, we can render clouds under environment illumination while having trained only for simpler, purely directional illumination conditions.

## 5 TRAINING

We train, validate, and test our networks on a mix of 80 procedurally generated, simulated, and artist-created clouds[2] represented by voxel grids with resolutions within $[100\text{–}1200]^3$. The geometry of each cloud is uniformly scaled to fit into a unit box and the density scaled accordingly to preserve the look. For testing, we carefully handpicked 5 clouds to cover the spectrum of shapes and thicknesses; these clouds are not used for training or validation. The remaining 75 clouds were used to generate training and validation data.

### 5.1 Data Generation

To compute a single training/validation record $(\mathbf{z}_i, L_i(\mathbf{x}_i, \omega_i))$, we sample a configuration $\mathcal{S}_i$, estimate the indirect in-scattered radiance $L_i(\mathbf{x}_i, \omega_i)$ thereof, and extract the descriptor $\mathbf{z}_i$. The shading location $(\mathbf{x}_i, \omega_i)$ is picked by first sampling a random direction $\omega_i$ uniformly over the solid sphere. Next, a ray origin $\mathbf{x}$ is picked such that the cloud's bounding sphere's cross section is hit with a uniform probability density by the ray parametrized by $(\mathbf{x}, \omega_i)$. Lastly, we trace the ray into the cloud by sampling the free-flight distance to obtain point $\mathbf{x}_i$. If this step does not produce a position within the cloud, we repeat the entire process. This procedure ensures that the density of records is higher near the cloud "surface", mimicking the distribution of records as found when path tracing, as is desired for the rendering scenario described in Section 4.4.

In typical scenarios, clouds are illuminated primarily by the Sun and the environment. Since the Sun can be parameterized well by a single direction, and the illumination from the sky and the ground can be approximated as a superposition of multiple directional lights, we train only for directional light sources. We sample a random direction $\omega_l$, from which light will arrive at the shading location, and then estimate the indirect in-scattered radiance $L_i(\mathbf{x}_i, \omega_i)$ using standard MC path tracing with next-event estimation; we draw samples progressively until the sample mean lies within a ±2% interval around the true answer (estimated with 95% confidence).

To study the impact of the phase function, we generated two data sets, each with $N = 15$ million samples. The first data set uses the tabulated chopped Lorenz-Mie phase function and reduced density; the version *with* the peak is used only at the first bounce while rendering. The second set uses a Henyey-Greenstein phase function with $g = 0.857$, matching the mean cosine of the Lorenz-Mie phase function with the peak.

### 5.2 Training Procedure

We use the squared error loss function between the logarithm of the network output and the $L_i(\mathbf{x}, \omega)$ estimated while generating data. Specifically, for a minibatch $\mathcal{B}$ the loss is

$$\ell_{\mathcal{B}} = \frac{1}{|\mathcal{B}|} \sum_{i \in \mathcal{B}} \left( \log\left(1 + g(\mathbf{z}_i; \boldsymbol{\theta})\right) - \log\left(1 + L_i(\mathbf{x}_i, \omega_i)\right)\right)^2.$$

We solve Equation (2) using stochastic gradient descent with the Adam update rule using the default learning rate, the decay parameters reported by Kingma and Ba [2014], and minibatches of size $|\mathcal{B}| = 1000$.

Each of the two data sets described in Section 5.1 is split 75% : 25% between non-overlapping training and validation subsets. The first is used for training the parameters of the network and the second to monitor convergence and select hyperparameters.

[1] We keep $\mathbf{W}_1^1$ only for equation brevity; it could be excluded since $\mathbf{o}^0 = 0$.
[2] A subset is available on https://www.disneyresearch.com/publication/deep-scattering.

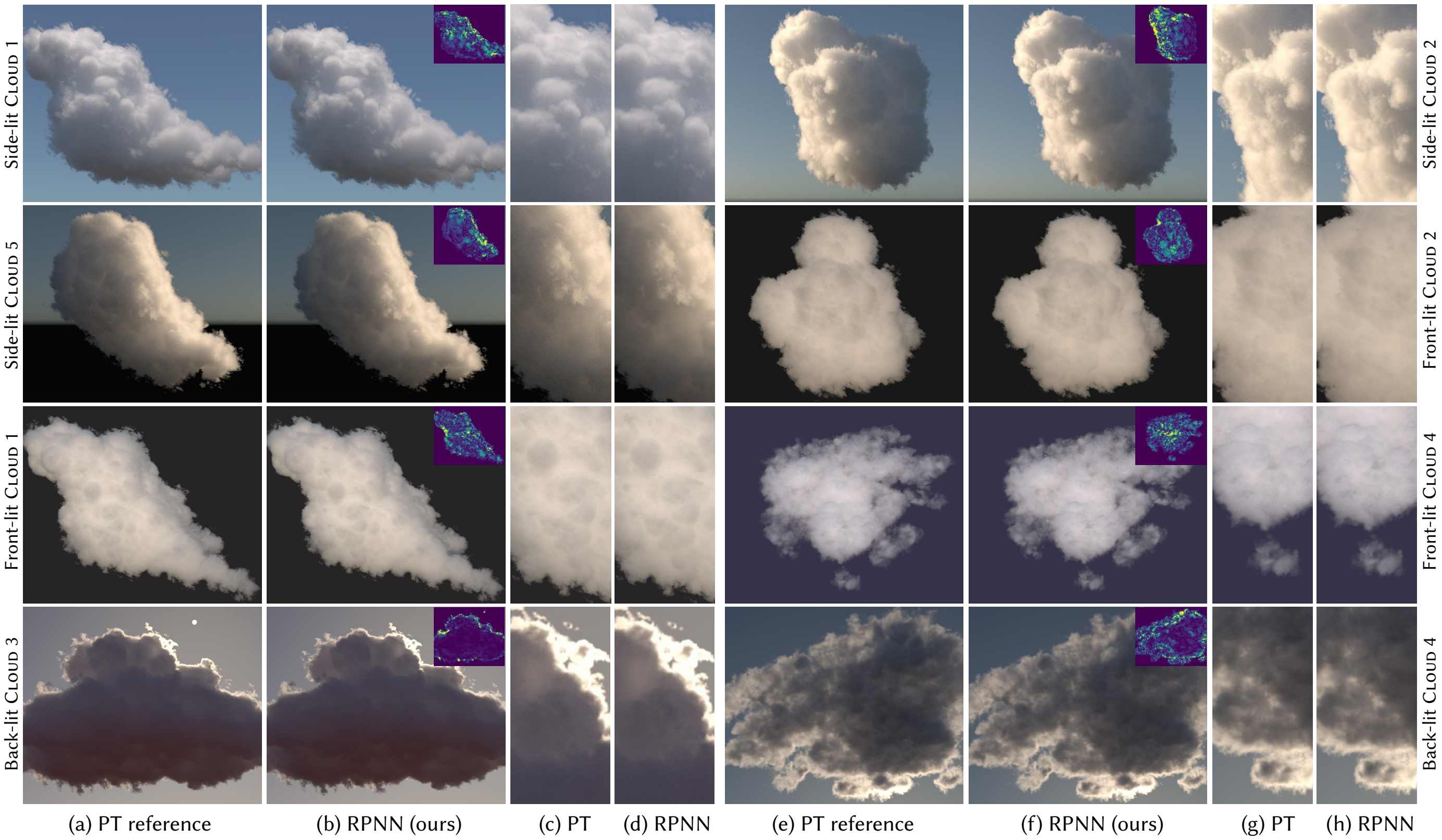

Fig. 7. Comparison of path-traced references and our approach that employs radiance-predicting neural networks (RPNN) on several different clouds illuminated by a sun-sky model. Our images accurately predict the incident illumination both due to the sun and the sky, which is estimated by sampling a number of light directions and evaluating our predictive model on them. When considering only sun-sky illumination, our method converges 24× faster than path tracing on average. Please see the supplementary material for additional results.

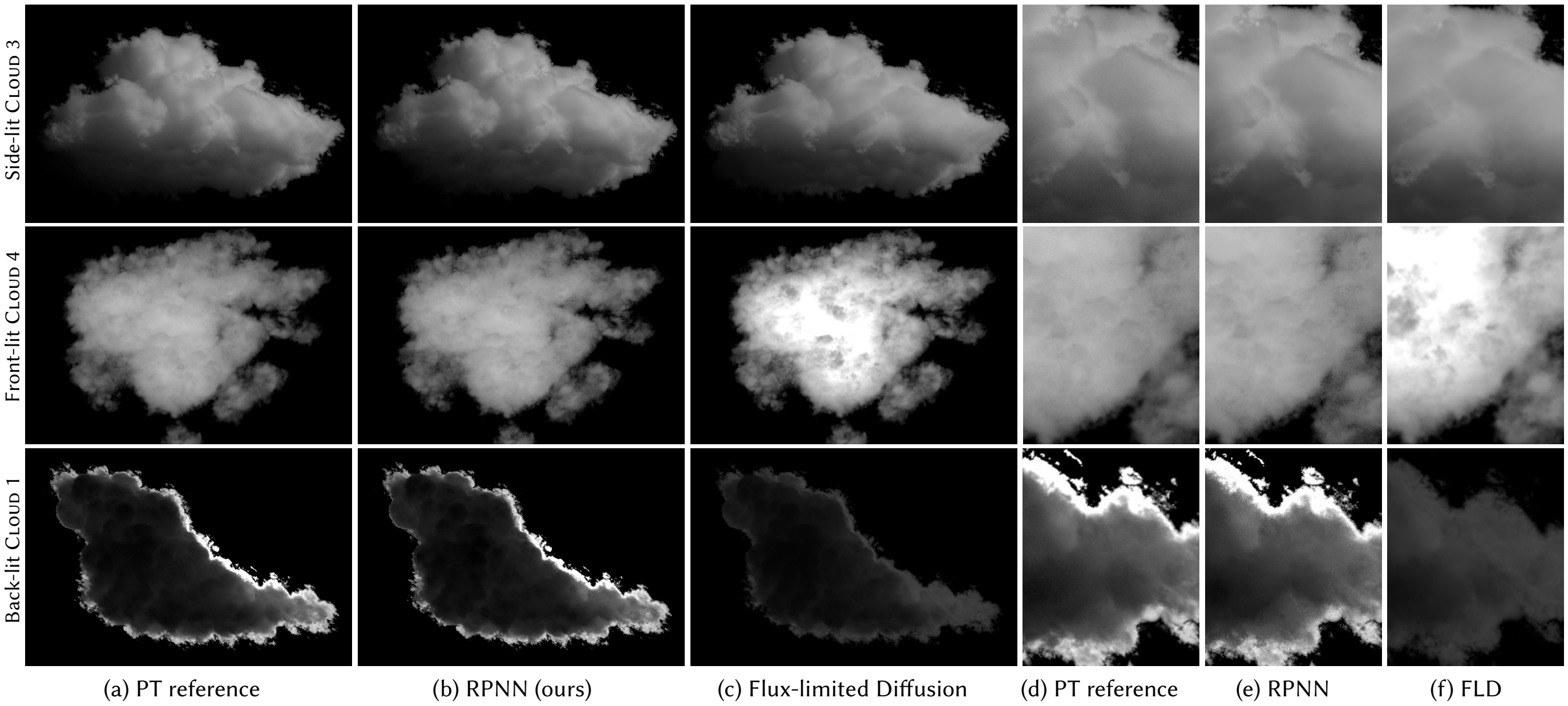

Fig. 8. Comparison of path-traced references, our method, and flux-limited diffusion (FLD) on three different clouds illuminated from three distinct directions. In contrast to FLD, our data-driven method correctly captures subtle contrast details (top), better preserves overall brightness of thin clouds (middle), and produces effects such as silverlining when trained on simulations of Lorenz-Mie scattering that FLD approximates by isotropic scattering. Columns on the right show zoom-ins. All FLD and RPNN images were rendered in approximately 6 seconds.

## 6 RESULTS

We compare the results obtained with our data-driven method to reference images rendered with a volumetric path tracer, and approximate solutions obtained with flux-limited diffusion (FLD) [Koerner et al. 2014]; we use the authors' GPU implementation. The five clouds shown in the figures here and in the supplemental materials were excluded from the training and validation sets.

*Implementation.* We implemented our prediction model in TensorFlow [Abadi et al. 2015], including the camera-ray generation, free-path sampling along primary rays, and single scattering estimation to allow for an efficient integration and execution on the GPU. For evaluating the densities, we constructed a Gaussian-filtered density mipmap of the cloud to avoid aliasing artifacts and flickering in animated sequences. We used $\sigma_i = 2^i$, where $i$ is the mip level, as the standard deviation of the Gaussian filters. All our and FLD results were obtained using an NVIDIA Titan X (Pascal) GPU. The path-traced reference images were computed on two Intel Xeon E5-2680v3 CPUs (24 cores / 48 threads in total). We also experimented with GPU path tracing, which, however, was close to 2× slower than the dual-CPU setup due to the incoherent memory accesses inherent to tracing long light paths through clouds. We therefore compare our method to the faster CPU baseline. All images were rendered at 1280 × 720 resolution. The rendering time depends linearly on the number of pixels and can be controlled by adjusting the resolution.

*Comparison to Reference.* Figure 7 compares our solution to a path-traced reference on a number of different clouds in varying lighting configurations. All images in the figure use a sun-sky model; see the supplementary materials for results with HDR environment maps. In most cases, our neural network predicts the radiance values accurately, producing images that are nearly indistinguishable from the reference; we recommend studying the differences using the supplemented interactive viewer. The highest errors appear in the back-lit Cloud 4 (bottom right), where the network produces an image with slightly less contrast. This is partly due to the fact that our data set of 80 clouds consists mostly of cumulus clouds, whereas the problematic cloud is closer in shape to a stratus.

Since both methods are Monte Carlo estimators, we can quantify their efficiency in reducing noise using *time to unit variance* (TTUV), defined per image as the product of render time and mean pixel variance. The TTUV can also be interpreted as the variance after rendering for one unit of time. Computing the ratio of the TTUV of the reference and the TTUV of our method yields the speedup we achieve over path tracing, which is reported alongside the incurred bias (of converged results) in Table 1. Our approach is up to three orders of magnitude faster than the path tracer under directional illumination, meaning that a high-quality image produced by the path tracer within a day can be synthesized using our method in mere seconds. Having to integrate numerically over multiple light directions reduces our speedup since much of the noise comes from the directional sampling. Hence the sun-sky lighting and the relatively low-dynamic-range envmap render generally slower; we discuss a possible remedy in Section 7. In the case of directional illumination, the speedups are lower in back-lit configurations, where most energy is contributed by single scattering that we do not accelerate.

Table 1. Speedup over path tracing and bias of converged images computed with our technique for 5 different clouds under directional illumination, a sun-sky model, and environment maps with medium dynamic range. For directional light sources we compare against flux-limited diffusion [Koerner et al. 2014] and achieve one to two orders of magnitude lower bias.

| | | Bias FLD | Bias RPNN (Ours) | | | Speedup RPNN (Ours) | | |
|---|---|---|---|---|---|---|---|---|
| Scene | Light dir. | Dir. illum | Dir. illum | Sun-sky | Envmap | Dir. illum. | Sun-sky | Envmap |
| Cloud 1 | Side | 3.36e-02 | 2.32e-03 | 2.02e-03 | 1.21e-03 | 4000 × | 41.7 × | 19.1 × |
| | Front | 2.61e-02 | 1.73e-03 | 2.26e-03 | 1.54e-03 | 1903 × | 26.9 × | 18.5 × |
| | Back | 1.92e-01 | 4.87e-03 | 1.57e-02 | 5.46e-04 | 70 × | 10.0 × | 4.7 × |
| Cloud 2 | Side | 1.42e-02 | 1.88e-03 | 4.01e-03 | 1.13e-03 | 3006 × | 18.7 × | 10.6 × |
| | Front | 2.38e-02 | 1.53e-03 | 2.96e-03 | 1.32e-03 | 685 × | 28.7 × | 5.9 × |
| | Back | 9.69e-02 | 5.15e-03 | 5.18e-03 | 1.87e-02 | 38 × | 13.0 × | 3.6 × |
| Cloud 3 | Side | 1.28e-02 | 2.55e-03 | 2.38e-03 | 1.31e-03 | 2727 × | 19.7 × | 6.3 × |
| | Front | 9.82e-03 | 1.54e-03 | 1.43e-03 | 1.46e-03 | 2532 × | 13.6 × | 19.1 × |
| | Back | 1.49e-01 | 5.24e-03 | 1.63e-02 | 7.69e-03 | 62 × | 5.3 × | 3.0 × |
| Cloud 4 | Side | 8.25e-02 | 1.81e-03 | 2.04e-03 | 1.24e-03 | 1308 × | 25.9 × | 12.1 × |
| | Front | 2.69e-02 | 1.75e-03 | 1.58e-03 | 1.70e-03 | 1282 × | 15.8 × | 9.9 × |
| | Back | 2.51e-01 | 6.35e-03 | 4.94e-03 | 6.06e-03 | 69 × | 15.5 × | 31.2 × |
| Cloud 5 | Side | 1.21e-02 | 2.08e-03 | 8.41e-03 | 1.61e-03 | 1964 × | 25.0 × | 9.8 × |
| | Front | 2.57e-02 | 1.32e-03 | 3.95e-03 | 1.16e-03 | 795 × | 36.6 × | 4.9 × |
| | Back | 1.68e-01 | 5.59e-03 | 1.47e-02 | 5.18e-04 | 30 × | 2.3 × | 2.9 × |

*Comparison to Flux-limited Diffusion.* In Figure 8, we compare also to FLD. Unlike our method, the diffusion theory-based FLD simulates all orders of scattering—including single scattering—with the inherent assumption of isotropic scattering. This prevents correctly synthesizing the silverlining effect, which stems from low-order, highly forward Lorenz-Mie scattering. While thin clouds are problematic for our method (due to a rather small number of those in the training set), FLD produces even worse results, overestimating the cloud brightness (middle row) and significantly changing the visual look of the cloud (see the supplementary material). In Table 1, we report the root-mean-square error on converged images of FLD and our method as bias. Our method outperforms FLD by one to two orders of magnitude in this regard.

*Impact of Phase Function.* In Figure 9, we compare renderings with Lorenz-Mie and Henyey-Greenstein phase functions synthesized by path tracing and our method. Being an approximation, our method does not match the path tracer exactly, however, the visual difference is smaller than between the two phase functions. Previous methods for fast rendering of clouds heavily approximated or ignored either the cloud geometry, or the shape of Lorenz-Mie phase function. Our method is the first to produce high-quality results within minutes.

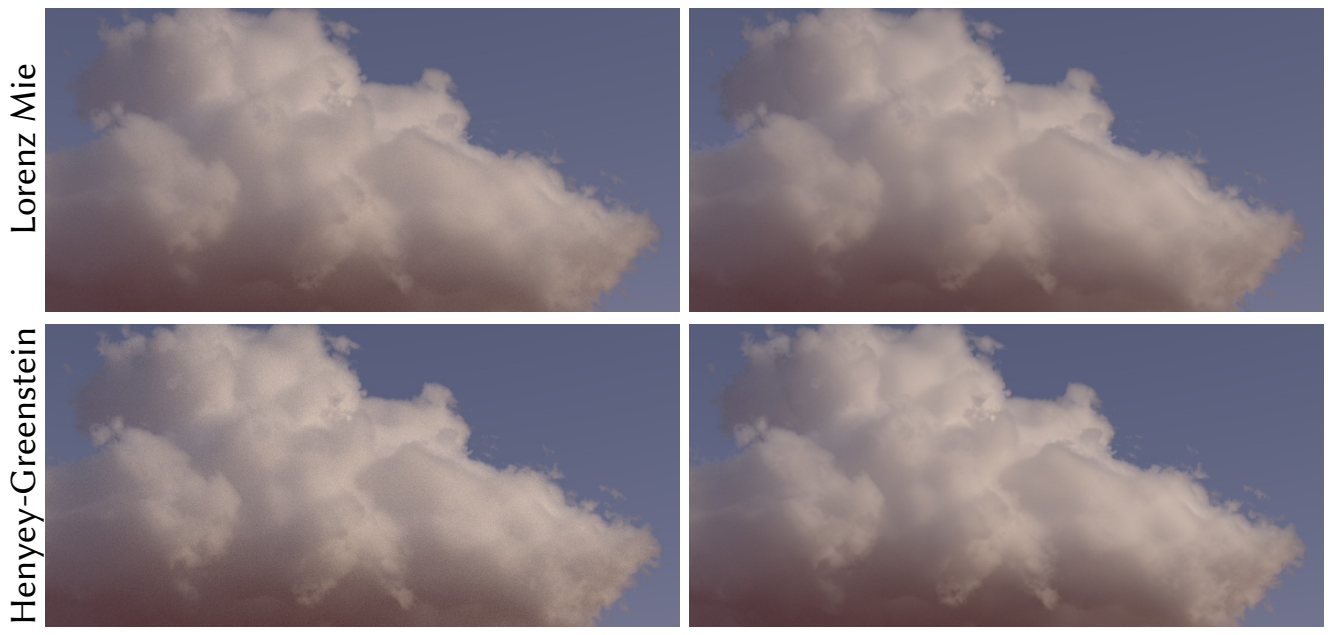

(a) PT reference (b) RPNN

Fig. 9. Visual comparison of Lorenz-Mie scattering and the Henyey-Greenstein approximation rendered with path tracing and our method trained on data simulated with the corresponding phase function. Note that the (vertical) difference between the two phase functions is larger than the (horizontal) difference between the reference and our solution.

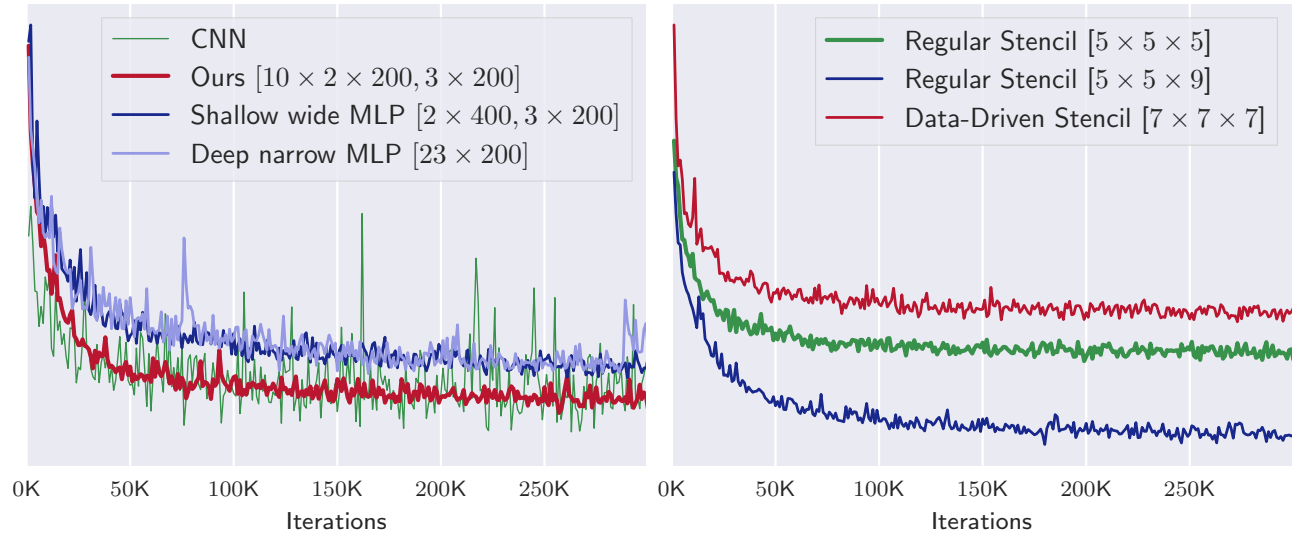

Fig. 10. Convergence of validation error for different network designs (left) and various stencil shapes with our network architecture (right).

## 7 DISCUSSION

In this section, we compare our network design to three other architectures that we tested, analyze the impact of certain hyper-parameters, and discuss limitations of our current approach.

*Comparison to Vanilla MLP and CNN.* We experimented with a vanilla MLP with different numbers of layers and nodes. Figure 10 compares our approach to a *shallow wide* MLP with 5 layers containing $[400, 400, 200, 200, 200]$ nodes, and to a *deep narrow* MLP with 23 layers each containing 200 nodes. The deep narrow MLP has the same number of layers as our design: we use 10 blocks, each with 2 layers, and 3 layers at the end (each layer contains 200 nodes). The main difference to our design is that we feed the hierarchical descriptor to the other two variants all at once; the entire hierarchy is input to the *first* layer. The other difference is the presence of residual connections in our design. All compared MLPs use about 1.4 million trainable parameters. As demonstrated by the convergence curves, our architecture learns faster and converges to a lower error. This highlights the benefit of the progressive feeding that provides means to better adapt to signals at different frequency scales.

We also tested a CNN; please see the supplementary document for the description of the architecture. We did not see quality improvements over our approach while incurring a significant performance- and stability hit. This may be because the size of the input is relatively small compared to applications where CNNs excel. Furthermore, the spatial structure which CNNs are designed for is fully captured by the input stencil. We conjecture that if the descriptor was sufficiently larger, an appropriately designed CNN might yield a quality improvement over our network at the expense of additional time spent constructing descriptors, which is already the computational bottleneck of our approach.

*Stencil Size.* We tested several stencils and their sizes. Figure 10 compares the convergence of the validation error with stencils of varying resolution and a data-driven stencil with points distributed according to the mean fluence distribution around the shading location; please see the supplementary material for details. Contrary to our intuition, the fluence-based approach never exceeded the performance of the straightforward regular stencils. We chose the $5 \times 5 \times 9$ configuration as it strikes a good balance between accuracy and the cost of querying the density values and number of trainable parameters in the network.

*Including Direct In-scattering in Predictions.* We compared the convergence error of learning *all* in-scattered radiance $L_s$ (green curve) and only the *indirect* component $L_i$ (blue curve). The network struggles with learning the full $L_s$ due to the very high-frequency nature of direct illumination. This component is largely responsible for the prominent silverlining effect in backlit configurations, and inaccurate predictions would be detrimental to the overall visual quality. Since the variance of integrating the direct in-scattered radiance via Monte Carlo is relatively low, excluding $L_d$ from the prediction helps attain high quality at tractable extra cost as shadow rays can be made coherent.

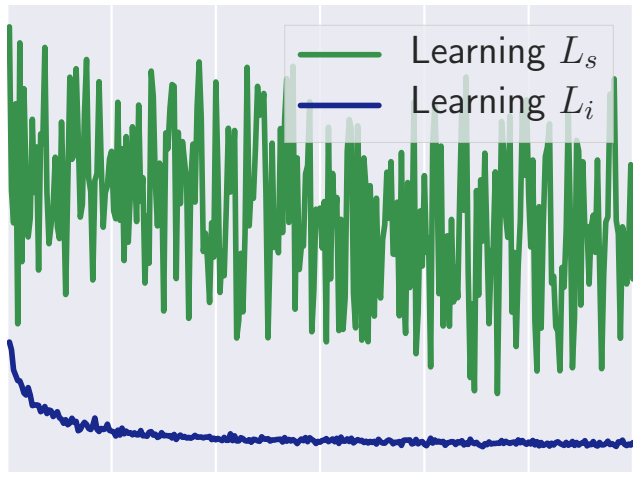

*Non-Cloud Shaped Objects.* We tried to predict radiance on non-cloud shaped objects using one of our models we trained on clouds (see Figure 11 and the supplementary material). As expected, our model performs worse than on clouds, given that the training data is not representative of the rendered shapes. While general characteristics of light transport are present, the brightness of subsurface scattering is overestimated. These experiments highlight another limitation of our approach: our stencil—being unable to resolve detail at large distances—cannot capture distant hard shadows well. We do not expect a proper training set to fully alleviate this problem.

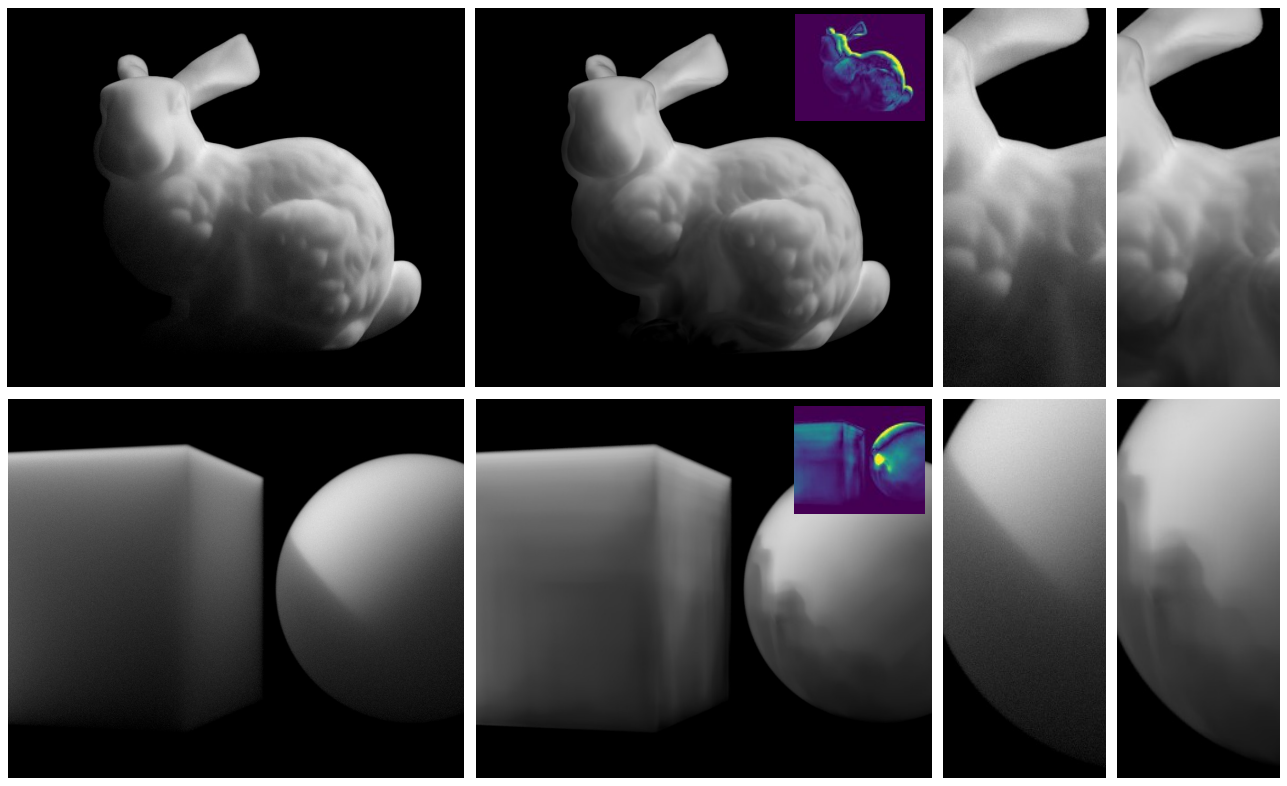

(a) PT reference (b) RPNN (Ours) (c) PT (d) RPNN

Fig. 11. Comparison of path-traced references and our method on the StanfordBunny and the SphereBox scenes under directional illumination. On such shapes, which are radically different from the clouds that the network was trained on, the incurred bias is substantially higher. Furthermore, due to the coarse approximation of distant shapes by our stencil, hard shadows can not be well synthesized (bottom row). Please see the supplementary material for additional results.

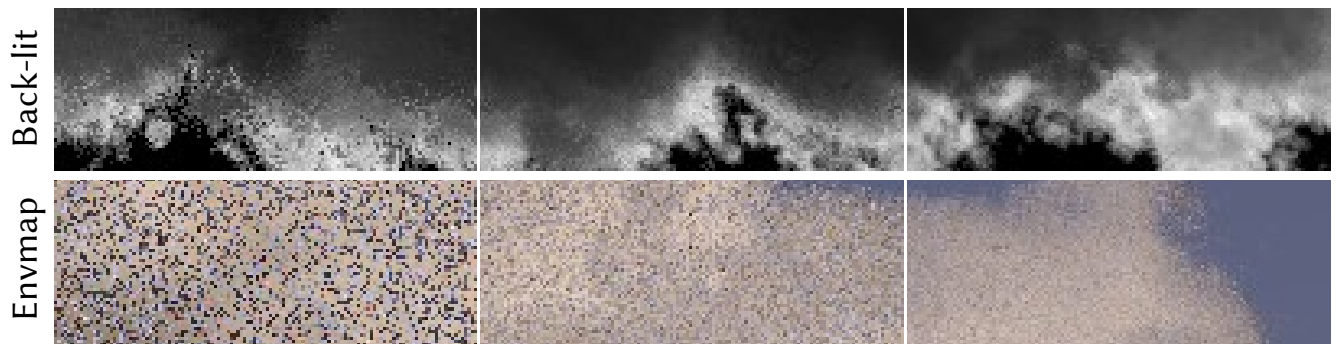

Fig. 12. If rendered at low samples per pixel, images may suffer from noise due to the MC integration that is used for sampling visible points and estimating single scattering. The noise level increases when lighting the cloud by the surrounding environment (bottom). Our method requires 1-3 seconds per SPP for images in our dataset.

*Environment Lighting.* Since we estimate part of the transport using MC integration (distance sampling and single scattering), the rendered images may still suffer from a small amount of residual noise. Figure 12 compares the levels of noise with different numbers of samples per pixel (SPP). The noise is exacerbated by environmental lighting, which requires multiple samples/predictions to cover the full sphere of directions. An interesting solution to avoid sampling the environment map would be to expand it into a suitable basis, e.g. spherical harmonics, and feed the expansion coefficient into the network, thereby training it to predict the radiance due to the full sky and ground. While fruitful, we believe this topic warrants its own investigation and leave it for future work. The noise could also be removed by applying an a-posteriori denoiser.

*Retraining and Adaptation.* Being data-driven, the performance of our method depends largely on how well the training set represents the actual rendering scenario. That being said, our networks require ~12 h of training on a single GPU. Tailoring their performance to a particular kind of cloud or specific cloud formations is thus relatively easy, provided that enough training data is available. Lightweight retraining could also be achieved by employing the idea of progressive networks [Rusu et al. 2016].

*Temporal Stability.* A key strength of our approach is its temporal stability stemming from the hierarchical descriptor and filtering. The accompanying video shows sequences with an orbiting camera and light sources, a cloud with changing density, and a cloud with changing shape, all of which are temporally stable.

## 8 CONCLUSION

We presented a novel approach for rendering atmospheric clouds at fast rates. Our radiance-predicting neural networks represent a new point in the spectrum of methods trading accuracy for speed. We achieve results that are nearly indistinguishable from the reference often faster or at comparable cost to previous approaches. The key and novel ingredient of our method is the progressive feeding of a hierarchical descriptor into the network. We further improve performance using standard techniques, such as residual connections and mip-mapping. Our method is temporally stable and could present a viable solution for applications such as asset design, future computer games, but also offline rendering if a small amount of bias is visually acceptable. We demonstrated that deep neural networks can be successfully applied to transport problems and hope that our work will stimulate further developments in that direction.

## ACKNOWLEDGMENTS

We thank Delio Vicini for proofreading and helpful discussions, the Stanford 3D Scanning Repository for the Armadillo, Dragon, and StanfordBunny models, and Magnus Wrenninge and the Pixar Animation Studios for providing most of the cloud volumes used throughout project. We are also grateful to David Koerner for sharing the source code of the flux-limited diffusion [Koerner et al. 2014].